# On the Management of Type 1 Diabetes Mellitus with IoT Devices and ML Techniques

Ignacio Rodríguez

**Abstract** - The purpose of this Conference is to present the main lines of base projects that are founded on research already begun in previous years. In this sense, this manuscript will present the main lines of research in Diabetes Mellitus type 1 and Machine Learning techniques in an Internet of Things environment, so that we can summarize the future lines to be developed as follows: data collection through biosensors, massive data processing in the cloud, interconnection of biodevices, local computing vs. cloud computing, and possibilities of machine learning techniques to predict blood glucose values, including both variable selection algorithms and predictive techniques.

## I. INTRODUCCION

In 2021, Diabetes Mellitus (DM) is undoubtedly one of the leading health problems in the world, especially in more developed countries. Factors associated with the western life, such as a sedentary lifestyle, obesity, lack of physical activity, poor eating habits, and others related to genetics or age are intimately connected with the development of this pathology.

Today, it is estimated that diabetes affects between 5% and 10% of the Spanish population, according to the studies consulted. In this sense, in the last National Health Survey in Spain, for year 2017, 7.8% of the entire population had been diagnosed with DM. Specifically, this figure rose to 21% in those over 65, while in 1993, it only reached 4.1% [1].

DM is characterized by a continuous elevation of blood glucose, either due to a lack of endogenous insulin production, or resistance to its action, referring to the former as Type 1 and the latter as Type 2 (DM1 and DM2, respectively). In both cases, the consequences of presenting abnormally high blood glucose levels, sustained over time, are devastating for the body.

Thus, DM is a multipathology, which means it generates multiple possible ailments related to the evolution of the disease. In this regard, it should be noted that DM is the primary cause of: retinopathy, blindness under 65 years of age, kidney transplantation, cardiovascular diseases, etc. Apart from the human drama suffered from these complications, DM generates a growing and enormous health expenditure worldwide, especially in Western societies, due to its characteristics, which are determined by an abuse of excessively processed foods calories, as well as the lack of physical activity.

The average annual direct health cost of each patient with DM exceeds the average cost of a person without this pathology by 2,145 euros [2]. Therefore, DM imposes an economic burden that could reach 2.5% of GDP, with an estimated 19,908,661 million euros in 2015. If only a few specific attitudinal changes are made in relation to some of the risk factors listed above, the company could save 64.8% of that cost, around 12,900.8 million euros (between 2,428.5 and 17,764.2 million). Of these possible savings, the greatest would be achieved through better diet control, which appears to be responsible for 40% of the incremental social costs of the disease.

However, the complications described above can be contained, delayed and even avoided altogether if the diabetic patient's level of control is increased. The Diabetes Control Complications trial (DCCT) [3] and the United Kingdom Prevention Diabetes Study (UKPDS) [4], for DM1 and DM2 respectively, show that good metabolic control is essential for the prevention of ailments associated with the course of diabetes.

In addition, these studies prove that the improvement in glycemic control is clear if the patient plans:

Food, according to the needs of the person.

Physical exercise, again according to the specific characteristics of the patient, age and general state of health.

Medication (insulin, hypoglycemic agents), following the instructions of the endocrine doctor.

Healthy habits, with adequate sleep and stress control.

Periodic checks, through the various options of existing glucose meters and through periodic medical checks.

These indications are already highlighting the variables which need to be considered when managing the evolution of the DM. The diabetic patient must keep all this information in mind, which he must process subjectively, and use this information to decide on adjustments to his routines and in his medication to try to control the course of his blood glucose, always with the collaboration and due advice of his healthcare professional. This control is singularly complex in the case of DM1 since, in this case, the pancreas stops producing insulin in the so-called beta cells of structures called the islets of Langerhans. This type of diabetes is characterized by its development in childhood or adolescence, and its origin is considered autoimmune. Insulin is the hormone needed to introduce blood glucose into cells that demand energy. Thus, the patient has to inject it exogenously, establishing the doses

---

Ignacio Rodríguez is with University of Málaga.

according to medical recommendations and correcting them, if appropriate, according to their own experience and, traditionally, from specific measurements of blood glucose with the use of test strips. In this manner, DM1 is the most aggressive type of DM due to its more pronounced glycemic oscillations, the absence of endogenous insulin that can buffer these fluctuations, and the health consequences that can occur in patients who suffer from it, both short term and long term.

In DM1, the subjective interpretation of the context by the individual means that, at times, errors in treatment may occur due to this partial assessment. However, new devices have been developed to provide more DM management options. The introduction of Continuous Glucose Meters (CGM) has led to a revolution in diabetes management. These instruments allow the continuous measurement of the blood glucose present in the subcutaneous interstitial fluid, with a sampling frequency normally between once per minute or once every five minutes such that, by means of correction algorithms, they present the blood glucose values. In this manner, the patient's knowledge of his blood glucose is uninterrupted. Figure 1 presents one of the most popular meters, the Abbot Freestyle Libre[1], which uses a Near Field Communication (NFC) monitoring system to show the user's glycemia on demand. Because of this feature, it is called Flash Glucose Monitoring (FGM).

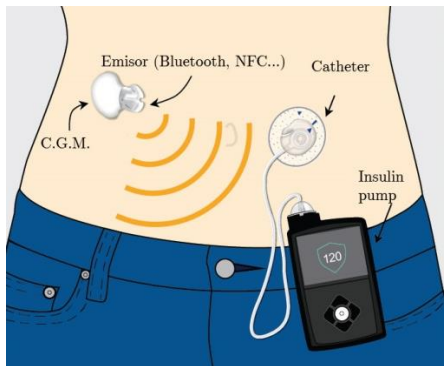

Figure 1. CGM Freestyle Libre    and Pump

Together with this, the DM1 patient has other equipment that can help manage his illness. Insulin pumps allow the patient to administer their dosage more comfortably and accurately, through a continuous infusion which, through a catheter, is absorbed under the skin (Figure 1).

The joint use of continuous sensors and pumps is due, in part, thanks to the improvement of short-range wireless communications, such as Bluetooth Low Energy, NFC (cited above), or even WiFi networks. Thus, a personal network (Body Area Network, BAN) can be generated, where the information will be managed. The use of smartphones, increasingly with connection capacity and computing power, provides all sorts of possibilities in this regard, making use of the potential that the Internet of Things (IoT) has been providing in recent years. This new paradigm indicates a promising research path to follow.

In addition, it should be noted that, until now, the management that takes advantage of the communications indicated in the previous paragraph is carried out only taking into account the evolution of glycemia, provided by the continuous meter. If, on the one hand, it seems clear that this is the main variable to consider, it is also known that there are numerous circumstances that affect the glycemic evolution of a DM1 patient. Food, physical exercise, daily routine, sleep, etc. have all been shown to be influential when it comes to affecting blood glucose levels. Although a few years ago these variables were obviated due to the difficulty of obtaining these levels as a continuous signal (and, subsequently, managing it), with the appearance of new portable biometric devices capable of recording measures such as heart rate 24 hours a day, as well as physical activity and many others, it now makes sense to record these variables and approach the study of their relevance to glycemic evolution. With the inclusion of the variables described above, the glycemia predication made by a certain Machine Learning algorithm (ML) can be improved, increasing its accuracy and thus being able to not only interrupt the infusion of insulin in episodes of hypoglycemia, but also anticipate situations of hyperglycemia and increase the dose when required. In this manner, the improvement in the glycemic control of the DM1 patient would be more than evident.

However, the inclusion of more variables that characterize the patient with diabetes undoubtedly entails an increase in the needs of computational hardware to manage such information. In this sense, the limitations of portable elements such as a smartphone or a Raspberry Pi can be a problem, so it is necessary to adjust the amount of data to be used, achieving a compromise between the speed of execution of the algorithms and the accuracy of the solution obtained. Resorting to opportunities such as Cloud Computing can be an appropriate strategy to resolve this limitation, but it should also be studied to what extent a small device can provide a solution "locally", even building the model "live" under certain circumstances that make this a necessity (lack of internet connection or its instability, remote rural areas, or circumstances in which the devices have to be offline: airplanes, conferences, etc.).

With all of the above, the main motivation of this doctoral thesis is the study of possible technological solutions to facilitate better management of the information in order to control DM1, focused on achieving a more reliable, fast and computationally affordable prediction of future blood glucose.

## II. OBJECTIVES

Having presented the motivation of this doctoral thesis and its central theme, the objectives to be met are listed, which have served as a guide for the development of this work:
- Study the current technological possibilities with respect to their application in the treatment of DM1. For this, a complete review of the literature must be carried out beforehand.
- Apply the concept of the IoT. The possibilities of interconnectivity and ubiquity promised by this new paradigm make its application to DM1 ideal.

---

[1] https://www.freestylelibre.co.uk/libre/

- Increase the available information. Until now, the characterization of the diabetic patient has been carried out by means of specific measurements of his blood glucose, which expresses as a single value the state of the patient at that time.
- Adapt the treatment of information. Added to the above, and as a consequence, it will be necessary to adapt the processing of the data, increased by the previous objective.
- Increase the set of available data. The need to have an adequate and reliable set of data to carry out the experimentation is identified. Therefore, the objective is to carry out a campaign to collect data among patients with type 1 diabetes who meet the minimum requirements for reliability.
- Carry out a treatment of the nature and transmission of temporary data series. The information collected will have a main characteristic: this data is linked to a time stamp, forming time series which, in turn, can be transmitted wirelessly to a new receiver.
- Select the most relevant variables in the evolution of blood glucose. A classification can be established according to the influence on the variations of blood glucose values.
- Make a prediction of future values of time series data. Having formed the characteristics of the diabetic patient and the pre-processed data set, the possibilities that different machine learning algorithms (ML) offer in order to achieve a prediction of blood glucose values at a certain horizon are going to be weighed by prediction, in a manner that allows the patient to anticipate the management of their blood glucose values.
- Learn the stress limits of the devices to be used in the management of the DM1. Information management and the execution of prediction algorithms are tasks with a certain degree of demand which raises doubts about the capacity and solvency of some devices or others. Comparative tests will be done to assess this issue.
- Clearly present ideas and concepts. As a cross-cutting objective, it is intended to increase the competences in terms of presentation of results, development of proposals and conclusions (including the writing of the scientific articles published and of this report, as well as the oral presentation of the works presented in conferences). In addition, the work attempts, as far as possible and within the appropriate forums, to also fulfill the purpose of increasing scientific dissemination in a clear manner, with a pedagogical vision.

### III. RESULTS

Within the framework of this doctoral thesis, numerous contributions have been made, all of them included in scientific articles or presented through written and oral communications to conferences. Not all have been fully collected in the compendium of publications that make up this document because some are focused on issues which, even if related, escape the scientific unit presented in this text. However, an exhaustive list of works carried out during the temporary period of development of this thesis is presented at the end of this document (Bibliography, Publications).

The majority of works developed are related to the management of Diabetes Mellitus; nevertheless, others have focused on information management in other contexts, or on Information and Communication Technologies (ICT) in other areas, taking advantage of the knowledge gained in this regard with DM1. In any case, the experience acquired in the elaboration of some works undoubtedly results in the best execution of the following.

In order to focus on the technological management challenges presented by DM1, the possibilities offered by the IoT in regard to remote situation management [5] were studied. Thus, the IoT paradigm makes it possible to provide user-centered services, with continuous monitoring of the person.

The application of the previous concept has allowed us to propose an intelligent management platform for DM1, integrating in this IoT environment various channels of information that direct the data collected from the person either to themselves, or to health professionals or family, so that they know at all times the comprehensive state of the patient.

A proper characterization of the patient is achieved by means of intelligent devices that will indicate the situation: smartphones, smartbands and other portable sensors [6]. This is the main idea set out in other works, where, in addition, a novel treatment is proposed for the variables considered in DM1. Works of other authors have already noted the prolonged action, but of less intensity, presented by insulin beyond its short-term action, referred to at the time as "insulin on-board". This doctoral thesis proposes the extension of this concept to other collected variables, such as exercise, sleep, or food, understood as a remnant action of very low intensity lasting over a time of several hours, with a normally negligible action in time, but with additive effects that must be taken into account in the presence of multiple doses of insulin over a period of time, exercise of greater intensity, or a lack of rest.

Likewise, the idea of a complete characterization of a person has been extrapolated to other areas, so that, in situations of violence or aggression, data such as GPS position, health status, etc., can be known in order to intervene and safeguard the safety of the victim [7].

Likewise, novel data collection has been carried out. Thus, over a period of 14 days, 25 patients with DM1 were monitored through a continuous glucose monitoring device (CGM), obtaining, in addition to their blood glucose, their insulin doses and amount of food ingested. Added to this, other characteristics such as heart rate, physical exercise, sleep and schedules were collected (through a smartband). This makes this set of data one of the most relevant that can be found in the literature, since those used previously suffer from complete monitoring, a sufficiently representative number of participants, a minimally acceptable extension of the essay or even incorporate subjective annotations by the patient in some variables [8].

The information collected from a diabetic patient is characterized by being associated with a time stamp, resulting in the set of samples taken in time series. Due to its specific characteristics, this type of information must be treated appropriately, and this has been especially addressed, among others, in the works [9] [10]. Likewise, said information collected by the corresponding set of sensors can, in turn, be transmitted wirelessly to other receivers in the form of electromagnetic waves. These waves propagate throughout an environment in which there can be a whole series of obstacles that make communication difficult. Therefore, in order to study the attenuation that these signals may suffer (in order to ensure the establishment of the connection), the phenomenon of diffraction has been studied as the main mechanism of wave energy loss [11] [12] [13].

All the aforementioned variables jointly influence glycemia, its values and its evolution over time. However, not all do so to the same extent, as some are more influential than others. To analyze this fact, there are algorithms for selecting variables, some being specific for time series data. Of these, the Sequential Input Selection Algorithm (SISAL) has been selected, which has made it possible to catalogue the influence of the variables, as well as the time they take to be influential [7]. This is decisive in being able to make a prediction of blood glucose values, since it is possible to predict what should be included in a predictive algorithm and thus form a set of variables is not overloaded with the work of forecasting future values. With this study, it has been observed that the most influential variable is insulin, and the last 105 minutes of data should be taken. The next most-influential variable would be food, with the last 2.91 hours being relevant and, finally, the blood glucose itself, with results from just over 4 hours. The rest of the variables studied (exercise, heart rate, sleep and schedule) complete the set of variables in that order of importance.

Once a management platform has been proposed and multiple forms of characterization of the diabetic person have been generated, as well as a gradation of their influence, a blood glucose prediction can be generated at different time horizons. This work has been carried out, first of all, in the simplest way possible: taking into account only the past blood glucose values in order to predict the following. This univariate approach aims to be the first step in a whole series of predictive strategies. In this first study with a single variable, the amount of past values that should be taken into account, as well as their influence on the accuracy of the prediction, has been discussed. In the same manner, the possibility of a lower sampling frequency (and, consequently, a lighter and easily treatable signal) has also been discussed, concluding that a compromise between the information load and accuracy is possible [14].

The prediction algorithms play a fundamental role here and, in this sense, three well-known algorithms have been compared: AutoRegressive Integrated Moving Average (ARIMA), Random Forest (RF), and Support Vector Machines (SVM). The possibilities of the latter had already been studied previously, with promising results [15]. The univariate models developed were able to predict glucose values in a 15-minute predictive horizon with an average error of only 15.43 mg/dL, using only 24 past values collected in a 6-hour period. Increasing the sampling frequency to include 72 values, the error decreased to 10.15 mg/dL. The study in question has concluded that RF is the algorithm which provides the greatest accuracy, in general. This study was carried out during a 3-month predoctoral stay at the Sapienza Università di Roma, in the Computer Science, "Antonio Ruberti" Automation and Management Division, under the close supervision of Professor Dr. Ioannnis Chatzigiannakis.

Other predictive algorithms have also been tested during the development of this thesis. Thus, also applied to the prediction of time series, the novel PROPHET library of the mathematical software R was successfully used, demonstrating that the addition of meteorological prediction could increase the accuracy of the prediction of the energy consumption of a dwelling [16] [17].

Continuing with the development of the thesis, it should be noted that the algorithms used can become very demanding from the computational point of view when performing the task of blood glucose prediction. As mentioned, it was proposed to evaluate the extent to which a prediction can be made locally, that is, in a small device that the person can take with them (for example, a smartphone). For this, stress tests have been done replicating the same ML techniques applied in [18] on three devices: on the one hand, a powerful server and, in the same manner, on a smartphone and a Raspberry Pi. The results have indicated that certain techniques are require less processing than others but, in any case, without minimum technical requirements, local execution would be unattainable. However, the execution of the SVM algorithm has proved to be feasible in a smartphone with average characteristics, and this fact is even clearer if strategies are made to lighten the time series, such as reducing the sampling frequency. The results indicate that it is possible to model and predict future blood glucose values on a smartphone with a prediction horizon of 15 minutes and an RMSE of 11.65 mg/dL in only 16.15 seconds, sampling the last 6 hours every 10 minutes and using the RF algorithm. With the Raspberry Pi, the computational effort is increased to 56.49 seconds in the same circumstances, but it can be improved to 34.89 seconds if SVM is used, in this case reaching an RMSE of 19.90 mg/dL. Therefore, it is concluded that practically real-time glycemic prediction is possible using small devices. The results of this study are indicated in [19].

In a transversal manner, it has been attempted, throughout the development of this doctoral thesis, to simultaneously acquire competencies in the pedagogical-informative field, essential requirements in transferring knowledge to society. Although this intention has been sought in all the works indicated above (with the limitations imposed by the scientific linguistic registry), in a special way it has materialized in a series of publications of a didactic nature in which issues such as laboratory work have been addressed (measurement of the speed of light [20]), as well as others related to the study of learning [21]. This undoubtedly contributed to the completion of the transveral and informative research training of the doctoral student.

## IV. CONCLUSIONS AND FUTURE WORK

The numerous advances achieved in ICT and, more specifically, in the IoT paradigm have great potential in terms of the amount of information that can be collected from a person. This is why the possibilities that are generated in the field of e-health are tremendously numerous. The management of this information to obtain knowledge can be done using another paradigm, the techniques of ML and automated data processing. In this work, the improvements that the application of these concepts can bring to the DM1 management field have been examined.

Thus, after a theoretical study, an IoT platform model has been proposed in which the inputs to be considered are established, as well as the technological possibilities for managing DM1 through this data collection [22]. Undoubtedly, the proposal reached establishes a framework in which the following works of this thesis are framed, as well as other future work.

New variables which have also been studied are derived from the inclusion of portable monitoring devices. The novel concept of the existence of certain 'on-board' variables [23], as an extension of that already described by other authors for insulin, is undoubtedly a contribution in the treatment of time series variables for the study of DM1, but also for other types of ailments, since the monitored results will also be associated with a timestamp, and may present remaining actions analogous to insulin and other variables described in this thesis.

After the cited framework, a monitoring campaign of 25 people with type 1 diabetes has been carried out, which has been a challenge for the number of participants over 14 days, as well as for the management of the devices used and the data processing. This data collection has been an achievement in itself.

With all the data collected, the variables described above were categorized. With this categorization, an establishment of a ranking of importance has been added, indicating which variables are the ones that most influence the evolution of blood glucose. Once the signals of the different variables are adequate, there is a need to study whether the new monitored signals are really relevant to characterize the condition of a diabetic patient. To this end, a specific variable selection method has been applied for temporary data structures called the Sequential Input Selection Algorithm (SISAL). The results of this work have indicated that, indeed, a gradation of influence can be established, considering levels of importance between variables.

Another contribution of this thesis has been carried out in the task of predicting future blood glucose values in DM1 patients. A comparative equality of conditions has been established for three known ML algorithms using only past blood glucose values, opening a helpful discussion on the most influential circumstances in regard to accuracy: the amount of past data collected, the sampling frequency and the predictive horizon [24] . A comparison like this has not been found to be published previously in the scientific literature.

Finally, it has also been relevant to evaluate in a practical and real manner the possibility of implementing the previous prediction on a device, considering two options: calculation on a remote server (cloud), or on a local device (smartphone, Rasperry Pi) [9]. The results are of great importance to the implementation of a real and portable control, by the patient, for DM1.

However, all the previous achievements have raised more questions that have escaped the limits of this thesis, waiting to be developed in future work, as indicated below.

First, it is worth noting that more characteristics of the people participating in our study phase have been collected and, although we have worked with these data to categorize their influence, they have not been used for the prediction of glycemia, because, so far, Univariate algorithms have worked well. As the first future work, it is planned to include other variables to check if accuracy is improved, without neglecting the corresponding analysis of the calculation effort that this entails. For this, in addition to the algorithms already studied, attempts will be made with others already employed in other works, such as the library of R mathematical software, called PROPHET [25].

With the data collected, and expanding the data set so that more patients are sampled, other issues could be addressed. Related to the concept of 'on-board', extended to physical exercise, numerous experiences of diabetics who report hypoglycemia beyond a time window immediately following the performance of physical activity have been referenced. It is intended to isolate these situations from the data set to study this phenomenon in more detail.

Likewise, focusing on the detection of hypoglycemias, a future objective is established to study the already known relationship between their occurrence and the associated heart rhythm variations, as well as the years of evolution of the patient and the quality of their control, as it is known that a greater amount of hypoglycemia leads to the gradual disappearance of the associated symptoms. In this manner, these variables can be studied to check at what time and under what control (good/bad) the symptoms of blood glucose drops in the heart rhythm disappear.

Similarly, the influence of lack of sleep on increased insulin resistance during the next day has also been reported, resulting in higher blood glucose levels. As in the previous concept, it is intended to isolate these situations to quantify this relationship. In fact, going further, it is feasible to establish daily averages per hour of blood glucose values and check the number of hours of slept during the previous night.

All these questions lead to a better understanding of DM1 thanks to delving deeper into the possibilities that an IoT environment and the application of ML algorithms can provide. With all this, it is proposed to feed the blood glucose prediction signal back into the system so that it enters a controller and closes the control loop. A closed-loop control system, with a continuous input of the monitoring variables, a glucose prediction output and, in turn, a feedback input that modifies the amount of insulin to be injected by the corresponding infusion pump would constitute, no doubt, an artificial pancreas that would, obviously, be the one - expected by millions of people - "technological cure" of DM1.